\documentclass[11pt]{article}
\usepackage{eamt24}
\usepackage{times}
\usepackage{url}
\usepackage{latexsym}
\usepackage[small,bf]{caption} 
\title{Exploring the Correlation between Human and Machine Evaluation of Simultaneous Speech Translation}

\author{Xiaoman Wang\\
  University of Leeds\\
  {\tt mlxwang@leeds.ac.uk}  \And
  Claudio Fantinuoli\\
  KUDO\\
  University of Mainz\\
  {\tt fantinuoli@uni-mainz.de}}

\date{}

\begin{document}
\maketitle
\begin{abstract}
Assessing the performance of interpreting services is a complex task, given the nuanced nature of spoken language translation, the strategies that interpreters apply, and the diverse expectations of users. The complexity of this task become even more pronounced when automated evaluation methods are applied. This is particularly true because interpreted texts exhibit less linearity between the source and target languages due to the strategies employed by the interpreter.

This study aims to assess the reliability of automatic metrics in evaluating simultaneous interpretations by analyzing their correlation with human evaluations. We focus on a particular feature of interpretation quality, namely translation accuracy or faithfulness. As a benchmark we use human assessments performed by language experts, and evaluate how well sentence embeddings and Large Language Models correlate with them. We quantify semantic similarity between the source and translated texts without relying on a reference translation. The results suggest GPT models, particularly GPT-3.5 with direct prompting, demonstrate the strongest correlation with human judgment in terms of semantic similarity between source and target texts, even when evaluating short textual segments. Additionally, the study reveals that the size of the context window has a notable impact on this correlation. 
\end{abstract}

\section{Introduction}

The assessment of interpreting quality is a common practice in both professional interpreting and academic contexts. The results of these evaluations offer valuable insights for a wide range of stakeholders, including interpreter's clients, users, practitioners, educators, certification bodies, and researchers~\cite{han_interpreting_2022}.

Assessing quality in interpretation is a complex endeavor. Quality is not only challenging to measure, but it manifests also an ``elusive nature''~\cite[p.~7]{becerra_quality_2013} making it difficult to define. The notion of quality in fact may vary from one user to another, introducing a substantial degree of subjectivity in determining what constitutes a good translation of speech. Furthermore, the criteria for quality are contingent upon the type of interpretation involved. For instance, in conference interpreting, the emphasis is generally on the quality of the interpreter's output, encompassing aspects such as content, language, and delivery. In contrast, within community settings like social and healthcare interpreting, interactional competencies and discourse management play a crucial role in determining what quality is~\cite{kalina_quality_2012}.

Traditionally, the assessment of interpreting performances has been carried out manually, a methodology that comes with its own set of pros and cons. On the positive side, human evaluations offer a holistic view of quality by taking into account various facets of the communication process, thereby delivering a more nuanced understanding of interpreting performance~\cite{pochhacker_quality_2002,becerra_quality_2013}. Conversely, manual assessment comes with its own set of challenges, including being labor-intensive, time-consuming and costly ~\cite{wu_assessing_2011}. Furthermore, the results often have limited generalizability due to either the restricted scope of the data sampled or the inherent complexities associated with evaluating spoken translation.

In light of the limitations, there has been a growing interest to apply automatic metrics to the evaluation of interpreting performances. While traditional statistical metrics like BLEU have shown limited efficacy in capturing translation quality from a user's perspective, the emergence of semantic vectors and pre-trained, large-scale generative language models has yielded promising results, especially in the domain of written translation~\cite{kocmi_large_2023}. The application of these metrics is gradually extending to the field of spoken translation as well~\cite{han_interpreting_2021}. However, it must be mentioned that orally translated texts possess certain characteristics that might restrict the efficacy of employing metrics designed for written texts. Interpreters, especially in the simultaneous modality, tend to alter the text more extensively than translators, modifying the structure and omitting parts deliberately as a strategy rather than a deficiency. For example, interpreters may omit part of the original when experiencing cognitive overload, when they cannot comprehend the original message, to name just a few~\cite{korpalOmissionSimultaneousInterpreting2012}. This non-linearity between source and target texts renders the task of automatic evaluation even more challenging.

The adoption of easy accessible and robust automatic evaluation in interpreting offers several potential applications that could benefit a wide range of stakeholders. Firstly, the ability to provide instant feedback to trainees and practitioners would enable them to quickly assess their performance and pinpoint areas for improvement, also in real-time, creating a faster feedback loop that could substantially accelerate autonomous skill development. Secondly, automatic evaluation might aid organizations in consistently and objectively monitoring the quality of their multilingual services. Thirdly, automatic metrics that correlate with human judgments might serve as a useful tool for the continuous evaluation of machine interpretation.

This study addresses two primary questions: First, is there an automatic metric that aligns closely with human judgment and can thus be used to automate the accuracy evaluation of spoken language translation? Second, do these metrics evaluate human-generated translations, machine-generated translations, or both more effectively?

The rest of the paper is organised as follows. In Section \ref{RelatedWork} we present an overview of research in the field of automatic evaluation of interpreting performances. In Section \ref{Methodology}, we illustrate our research methodology, our data and the experimental design. Section \ref{Dataset} describes the dataset created for this task. Section \ref{HumanEvaluation} describes the process for human evaluation of the translations while Section \ref{MachineEvaluation} delves on the process followed for the automatic evaluation. Section \ref{Results} presents the results. Section \ref{ethics} introduces some ethical implications. Finally, Section \ref{Conclusions} concludes the paper with some discussion and remarks.

\section{Related work}\label{RelatedWork}
The evaluation of translation quality and in particular of accuracy or information fidelity, i.e. the correspondence between source-language and target-language renditions, has traditionally differed between computer science, with its tradition of abundant use of automatic metrics, and translation and interpreting community, with its focus on manual evaluation as perceived by experts and users.

In computer science, evaluation metrics such as Bilingual Evaluation Understudy (BLEU)~\cite{papineni_bleu_2002}, National Institute of Standards and Technology (NIST)~\cite{doddington_automatic_2002}, Metric for Evaluation of Translation with Explicit Ordering (METEOR)~\cite{banerjee_meteor_2005}, and Translation Edit Rate (TER) have been foundational in establishing benchmarks for Machine Translation Quality Estimation (MTQE). BLEU and NIST emphasise n-gram precision, with NIST uniquely weighting distinct n-grams. METEOR integrates both recall and precision, while TER quantifies requisite edits for optimal translation alignment. However, recent scholarly discourse have suggested that these metrics, while valuable, may possess intrinsic limitations in encapsulating the multifaceted subtleties and overarching context of linguistic structures~\cite{fernandes_devil_2023}. This acknowledgement has precipitated the exploration of advanced, data-driven methodologies for MTQE without references. Neural networks, characterized by their bio-inspired architectures, emerge as a compelling alternative. These computational structures excel in managing voluminous datasets, discerning intricate patterns, and, crucially, accounting for the inherent complexities associated with linguistic phenomena.

In the field of neural network architectures, the potential of Recurrent Neural Networks (RNNs), Convolutional Neural Networks (CNNs), and the groundbreaking Transformer for semantic similarity computations has been explored. Of these, Transformer-based models like BERT and GPT have gained considerable academic traction due to their outstanding performance across numerous Natural Language Processing (NLP) tasks~\cite{wang_document-level_2023,kocmi_large_2023,clark_electra_2020,xenouleas_sumqe_2019,yang_xlnet_2019,brown_language_2020,hendy_how_2023,clark_what_2019,vaswani_attention_2017}. Central to these architectures is the concept of embeddings: dense vector representations that capture the semantic essence of words or textual segments. Within this high-dimensional space, vectors situated closely denote semantic relatedness. In translation evaluation, embeddings offer a mediating semantic layer, enabling comparisons between source and target linguistic structures. However, the embeddings landscape is complex. Models from the Universal Sentence Encoder Multilingual (USEM) to the Generative Pre-trained Transformer (GPT) produce embeddings with varied purposes. For example, USEM is geared towards retrieving semantically aligned entities, while GPT emphasizes generating context-rich linguistic constructs. These nuances highlight the need for researchers to thoughtfully choose models aligned with their specific research goals.

In interpreting studies, a subdomain of translation studies dedicated to oral translation, the traditional practice has been to assess accuracy manually, with or without references. Many of the evaluation methodologies are derived by written translation. For the assessment based on references, also known as intra-lingua assessment, the notion of "tertium comparisons" stands as a pivotal benchmark within a particular language~\cite{setton_syntacrobatics_2007}. Tracing the historical evolution of this methodology, \cite{carroll_experiment_1966} stands out as a foundational contributor. He experimented with lay people to ascertain accuracy in translations. Building on Carroll's scale, \cite{tiselius_revisiting_2009} refined the process, integrating references to spoken language and interpreting for intra-lingual assessment. These approaches, while receiving considerable acceptance from seasoned interpreters, are not without limitations. The impact of cognitive-linguistic factors can potentially alter evaluation results~\cite{han_accuracy_2021}. Moreover, intra-lingual evaluations face challenges in adapting to changing contexts and demographics and may lack a universally acknowledged point of reference~\cite{setton_syntacrobatics_2007}.

Academic discussions prioritise evaluation methods for gauging accuracy in inter-language interpreting. These methods range from error analysis, as seen in works by \cite{gerver_effects_1969} and \cite{gile_fidelity_1995} that identify translation inaccuracies, to propositional analysis, endorsed by researchers like \cite{mackintosh_relay_1983} and \cite{lee_simultaneous_1999,lee_speech_1999,lee_ear_2002}, which examines textual accuracy. However, these methods present challenges in addressing linguistic subtleties and differing interpretations. More recent research emphasizes grading rubrics, tracing back to \cite{carroll_experiment_1966}, which outline performance across competency tiers and are validated in multiple studies~\cite{han_investigating_2016,han_using_2017,nia_rasch-based_2019,wu_analytic_2013}. Yet, even with proven reliability, this rubric-based evaluation faces hurdles like the development and validation of rubric descriptors and evaluator inconsistencies.

A few studies have explored the effectiveness of various metrics in evaluating the translation quality or interpreting performances.  \cite{chung_automatic_2020}, for instance, pinpointed the strong alignment between human evaluations and scores determined by BLEU and METEOR for German-to-Korean translation. Subsequent studies by \cite{han_can_2021} and \cite{lu_automatic_2022} reinforced the merit of these automated tools. Han and Lu (2021) discerned that METEOR's sentence-level evaluations resonated more with human assessments than broader, text-level evaluations. Conversely, Lu and Han's (2022) exploration, fortified with the integration of the BERT model~\cite{devlin_bert_2019}, showcased substantial correlations between human and automated evaluations, underlining the potential of these metrics in assessing interpreting performances. A recent study by \cite{kocmi_large_2023} employed Large Language Models like GPT to evaluate translation quality across three language pairs, concluding that only models GPT3.5 and above possess the capability for such translation quality assessment.

While initial studies have underscored positive and moderate-to-strong correlations for MT metrics such as BLEU and METEOR, to the best of our knowledge no research has been conducted so far on the use of language models for reference-free interpreting assessment. Our study aims to fill this gap. 

\section{Data and methodology}\label{Methodology}

\subsection{Dataset}\label{Dataset}
The dataset used for the study consists of 12 original speeches in English translated into Spanish, each lasting approximately 5 minutes. These videos were curated from a broader selection of real-life contexts, including lectures, business presentations, live tutorials, and political addresses.\footnote{The dataset is available under the Creative Commons 4.0 License at \url{https://github.com/renawang26/Information_fidelity}}

Although the corpus size is inherently limited, in order to allow high quality human evaluation, the selection of videos was strategically designed to capture a spectrum of speech features. The speeches were distributed equally in terms of gender, with 6 from male speakers and 6 from female speakers. In addition, the accent of the speakers comprises both native and non-native speakers. The nature of the speeches is diversified into three categories: 4 corporate, 4 political, and 4 general presentations. The speeches comprise 3529 tokens. 

For the evaluation purpose, each video was simultaneously interpreted in two ways:

\begin{itemize}
\item {\bf Translation H}: Human professional interpreters were engaged. Three interpreters, native Spanish speakers, were involved, each responsible for translating four videos. Simultaneous interpreters were required to interpret the entire video (approximately 5 minutes) to preserve the contextual information essential for accurate interpretation. However, only 2 minutes of the videos were randomly selected for the evaluation.

\item{\bf Translation M}: Machine interpretation was carried out by the KUDO AI Speech Translator, the only system available for simultaneous translation available at the moment of writing \footnote{\url{www.kudo.ai}}.
\end{itemize}

All recordings were automatically transcribed, and the transcriptions were post-edited for accuracy by expert linguists proficient in both English and Spanish. The goal of this operation was to make sure that the transcripts did not contain errors of transcription. The transcriptions were manually aligned based on semantic units, a critical step due to the absence of formal punctuation commonly found in written texts. These segments are roughly comparable to sentences or smaller paragraphs. The average length of segments is 29.41 tokens.

\subsection{Human evaluation}\label{HumanEvaluation}

The human evaluation process was guided by the methodology proposed by \cite{fantinuoli_towards_2021}, which uses a Likert scale to assess two key features of interpretation: accuracy (ability of the translation to convey the meaning of the original) and intellegibility (ability of the translation of being understandable). The two dimensions reflect the main criteria at the core of the product-oriented approach to quality evaluation in Interpreting Studies~\cite{tiselius_revisiting_2009}. For the purposes of this study, however, the focus was exclusively on the feature of informativeness, i.e. accuracy, leaving the assessment of intellegibility and any other potential feature for future research. One of the advantages of this framework lies in its being user-centric and in line with the corpus-based evaluation already established in Interpreting Studies to assess the quality of human interpretation.

The human evaluation was conducted using a diverse group of 18 evaluators. This consisted of 9 professional interpreters and 9 bilingual individuals without any translation or interpreting experience. The goal was to capture a broad and unbiased evaluation of the translations, taking into account both professional expertise and everyday bilingual proficiency. Each evaluator was assigned 4 videos to evaluate. They were informed that the translations were transcriptions of oral simultaneous interpretations.  

For each speech, the raters were asked to assess on a six-point Likert scale first the intelligibility of the output (without a comparison with the source speech nor a comparison between the two outputs), then to evaluate the accuracy of the renditions by comparing each one to the source speech.

An important feature of the evaluation process was the anonymity of the translation sources. Evaluators were not informed whether the translations were produced by a human or by a machine. This was a deliberate step to prevent any evaluation bias, ensuring that the judgment was strictly based on the quality of the translation, irrespective of the producer.

It is important to point out that with a value of 0.0964 the interater agreement is low (``slight agreement'' on a Fleiss' Kappa scale). This aspect showcases the intrinsic complexity of objectively assessing spoken language translation due to different expectations by the evaluators about what constitutes accuracy. This is an insidious limitation of the evaluation of human interpretation (see Han 2022). While the low agreement between multiple raters is expected, it also limits the generalizability of our findings.

\subsection{Machine evaluation}\label{MachineEvaluation}

Our approach to the machine assessment of spoken language translation is based on the concept of semantic similarity leveraging sentence embeddings and large language model prompting techniques. 
The rationale behind using embeddings to measure semantic similarity is multifold, as it proffers a host of advantages, such as the provision of a continuous representation~\cite{mikolov_distributed_2013,pennington_glove_2014}, incorporation of contextual information~\cite{devlin_bert_2019,peters_deep_1802}, dimensionality reduction~\cite{roweis_nonlinear_2000}, applicability of transfer learning~\cite{howard_universal_2018,raffel_exploring_2020}, multilingual support~\cite{conneau_word_2017,wu_googles_2016}, interoperability~\cite{ruder_survey_2019,artetxe_robust_2018}, ease of use~\cite{radford_improving_2018}, and state-of-the-art performance~\cite{vaswani_attention_2017,brown_language_2020} in NLP tasks. The advantage of using sentence embeddings over MTQE models for assessing interpreting performance lies in the ability of embeddings to evaluate without the need for references. This approach contrasts with some MTQE models, which typically depend on references for quality assessments. Furthermore, sentence embeddings are adaptable across a broad spectrum of languages and text genres, offering a versatile solution for evaluating interpreting performance. This flexibility is beneficial across different domains and contexts, whereas MTQE models often necessitate more specific training data to achieve comparable effectiveness. The fundamental operation of this methodology involves vectorising each sentence in both the source and target texts. Essentially, this means mapping each sentence to its corresponding vectors of real numbers, thereby projecting them into a shared multi-dimensional space. This conversion of textual data into numerical format empowers the machine to elaborate the semantics of the texts effectively. The subsequent step involves the calculation of cosine similarity, which serves as a measure of the similarity between each language pair.

We employed three neural network models to carry out sentence embedding: all-MiniLM-L6-v2\footnote{\url{https://huggingface.co/sentence-transformers/all-MiniLM-L6-v2}}, Generative Pre-trained Transformer (GPT), and in particular GPT-Ada.  model\footnote{\url{https://platform.openai.com/docs/api-reference/embeddings}}, and Universal Sentence Encoder Multilingual\footnote{\url{https://tfhub.dev/google/universal-sentence-encoder-multilingual-large/3}} (USEM). By integrating the all-MiniLM-L6-v2 for its efficient, compact design suitable for multi-language applications, alongside GPT-Ada for their advanced generative capabilities and adaptability, and USEM for its extensive language coverage and cross-lingual semantic understanding, this strategy offers a balanced and comprehensive approach. The embeddings obatained werer vectorised sentences in both English and Spanish. The sentence embeddings computed with these models were later used to calculate the Cosine Similarity between the source texts and the translations. 

In addition to the models for computing word vectors, we tested another method to compute semantic similarity leveraging the prompt functionality of GPT3.5\footnote{\url{https://www.openai.com}}. The Large Language Model was assigned the task of assessing the semantic similarity between pairs of sentences, one in English and the other in Spanish, using a scale ranging from 1 to 5. An example prompt provided was: "Given the two sentences in English and Spanish, rate from 1 to 5 their similarity, where 1 is not similar and 5 very similar." 

\subsection{Computing correlations}

The human and automatic assessments were put together in an evaluation matrix. Pearson's correlation coefficients were leveraged to explore the relationships between human and machine evaluations. Specifically, the correlations between human judgments and the cosine similarities derived from the embeddings of segments from the source speech and their corresponding translations (Translation H and Translation M) generated by three models. were calculated. As stated before, the overarching aim was to probe the feasibility of achieving machine-human parity in the evaluation results. 

Since the neural network models chosen for this experiment have limited reliably in computing semantic vectors for long texts, we opted to establish a correlation between human and machine evaluations at the segment level. It is essential to emphasize that similarities values obtained from isolated segment pairs have intrinsic limitations since they are not able to consider accuracy across segments. Thus, the quantity of tokens utilized as context for sentence embeddings could potentially affect the model's contextual comprehension and, subsequently, the precision of semantic similarity assessments.

To take this into consideration, we examined the effect of ``window size'', i.e. the number of segments combined into a single one. For this purpose, we computed similarities for window sizes up to five segments. The systematic variation in window size aimed to shed light on how semantic similarity between human and machine evaluations could potentially be influenced by the availability of cross-segment context. 

\section{Results}\label{Results}

To analyse the data, we devised charts from three perspectives, including the comparison of correlation values among evaluation methods, a comparison between Translation H and Translation M, and correlation values based on window size. 

\begin{figure}[h!]
    \centering
    \includegraphics[width=0.45\textwidth]{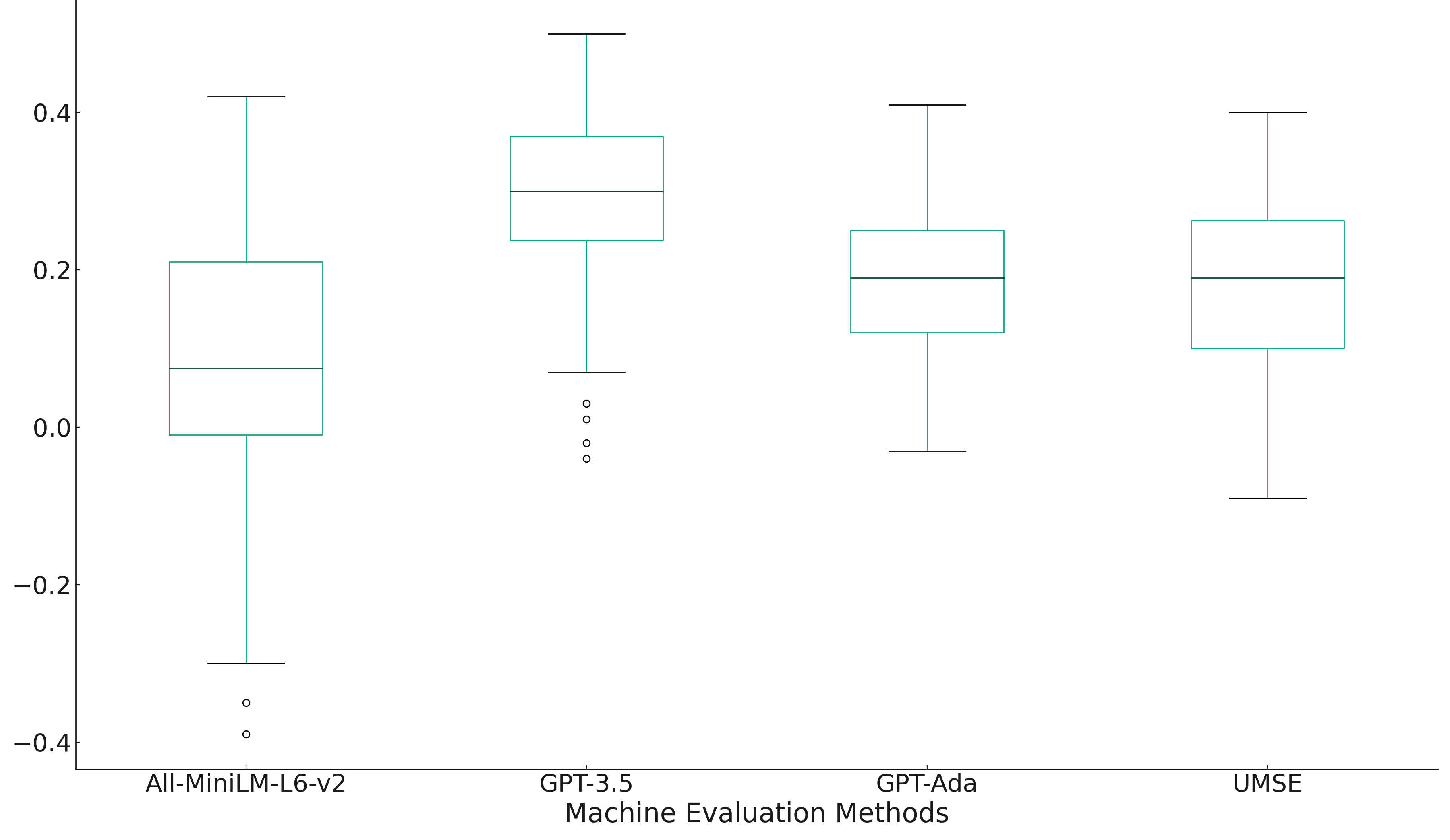}
    \caption{Correlations among machine evaluation methods}
    \label{fig:boxlplot}
\end{figure}

In Figure \ref{fig:boxlplot} we compare the distribution of correlation values across all machine evaluation methods, namely GPT-3.5, all-MiniLM-L6-v2, GPT-Ada, and UMSE, for both Translation H and Translation M. 

The correlation with GPT-3.5 displays the highest median correlation value. The interquartile range (IQR) is also quite narrow, indicating that the correlation values for this method are consistently high and well-aligned with human evaluations. The correlation with all-MiniLM-L6-v2 has a wide IQR, showcasing varied performance. The median value is close to zero, but there are negative outliers, indicating instances where the machine evaluation is inversely related to human judgment. The correlation values with GPT-Ada are relatively consistent, with a narrow IQR. The median is slightly above 0.3, which indicates moderate alignment with human judgments. UMSE’s performance seems to be close to GPT-Ada with a median slightly above 0.3. The IQR is a bit larger, suggesting a bit more variability in the correlation values.

For the comparison between Translation H and Translation M in Figure 2, paired bar charts elucidate the average correlation disparities for each machine evaluation method. GPT-3.5's measurements exhibit robust correlation values with human judgments for both translations, although Translation H marginally outperforms Translation M. For all-MiniLM-L6-v2, the correlation of Translation H gravitates towards zero, whereas Translation M registers a negative value, implying a potential misalignment of the all-MiniLM-L6-v2's evaluations with human perspectives, predominantly for Translation M. GPT-Ada embeddings yield nearly identical correlation values for both translations, but with Translation H slightly edging out. Intriguingly, UMSE's embeddings produce a higher correlation value for Translation M compared to Translation H.

\begin{figure}[h!]
    \centering
    \includegraphics[width=0.45\textwidth]{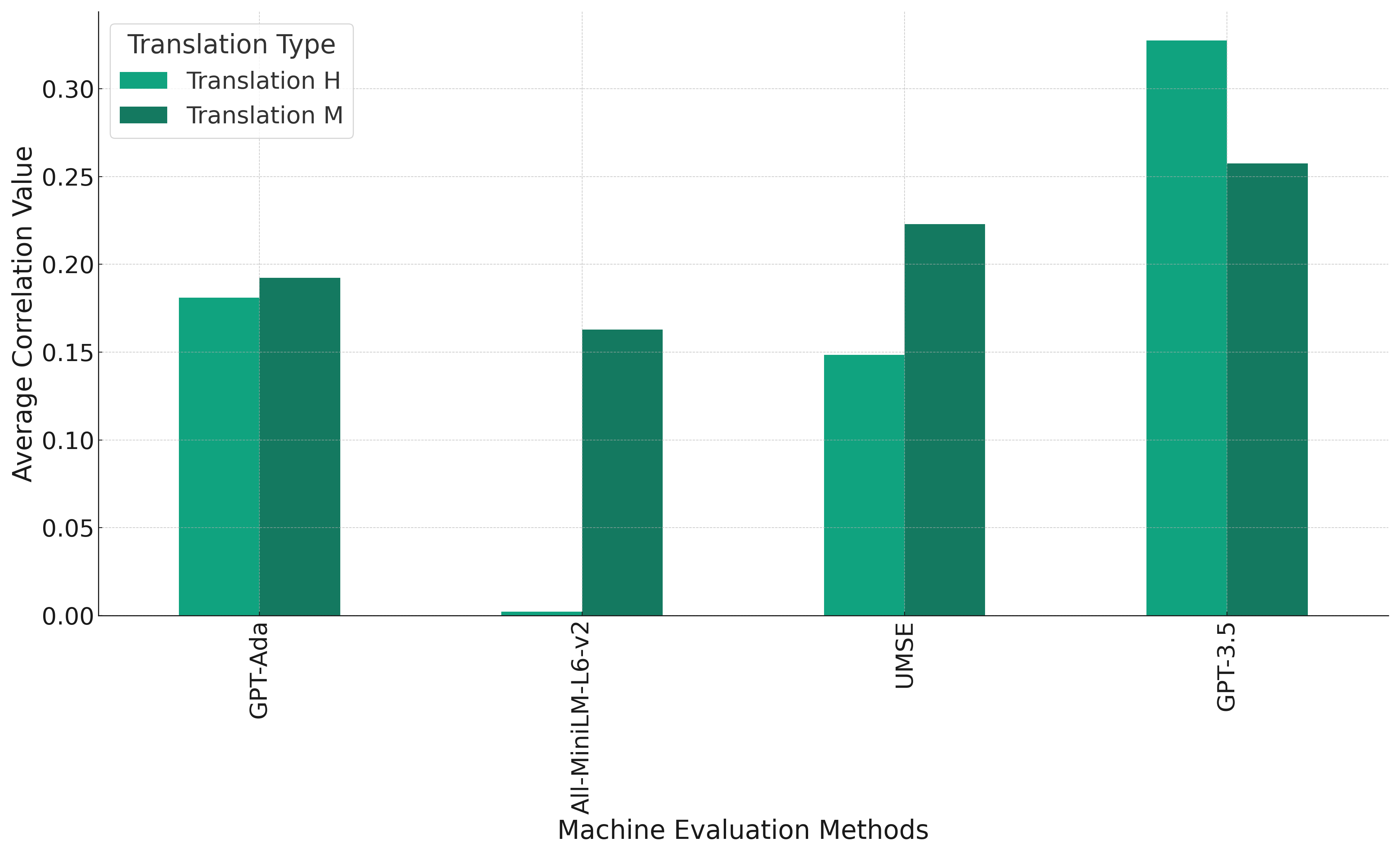}
    \caption{Correlations for Translation H and Translation M}
    \label{fig:bar_charts}
\end{figure}

Turning to the third perspective, which examines the shift in correlation values based on window size, line charts in Figure 3 and Figure 4 offer insights into this dynamic for each machine evaluation method. For Translation H (Human Translated) in Figure 3, GPT-Ada correlation with human ratings sees a mild fluctuation across window sizes, initially decreasing from window size-1 to size-2, then slightly rising in the following window size2-5. The all-MiniLM-L6-v2 model, in contrast, exhibits a downward trend, indicating reduced agreement with human evaluations as window size grows. UMSE consistently maintains a stable correlation with human ratings, showing only minor variations across different window sizes. GPT-3.5 presents a distinct pattern; while its correlation initially drops from size-1 to size-2, it surges notably in the subsequent window, outperforming the other models.

In observations for Translation M (Machine Translated) in Figure 4, GPT-Ada begins with a positive correlation with human ratings with size-1, but this declines as the window size expands, hinting at potential metric inconsistencies for broader contexts. The all-MiniLM-L6-v2 model's correlation, on the other hand, commences positively by size-1 and consistently rises with the window size, pointing to a more aligned evaluation with human judgment for larger translation segments. UMSE's performance mirrors its evaluation with human translations, maintaining stability across all window sizes and showcasing its consistent metric evaluation. In contrast, GPT-3.5's correlation fluctuates considerably across window sizes — experiencing a drop, a subsequent rise, and then another decline — indicating a variable level of concurrence with human assessments depending on the window size.

\begin{figure}
    \centering
    \includegraphics[width = 0.5\textwidth]{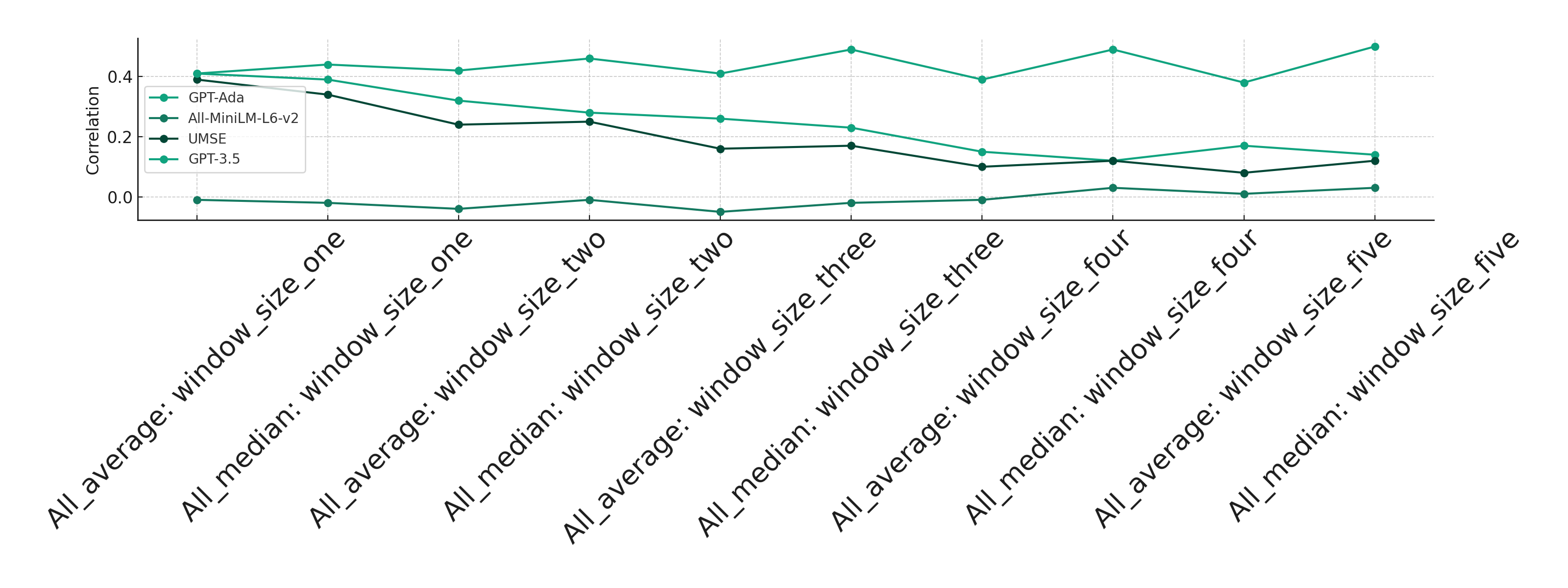}
    \caption{Correlations for Translation H according to window size}
    \label{fig:window_size}
\end{figure}

\begin{figure}
    \centering
    \includegraphics[width = 0.5\textwidth]{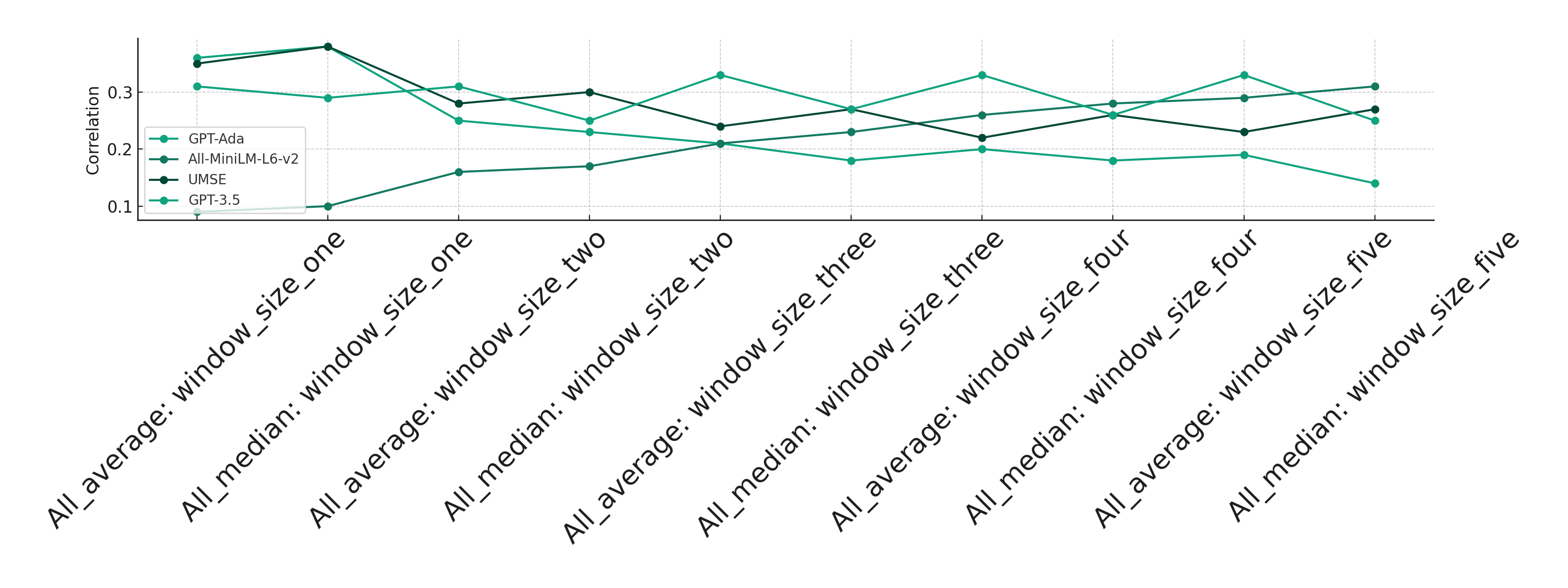}
    \caption{Correlation for Translation M according to window size}
    \label{fig:window_size2}
\end{figure}

\section{Ethical considerations}\label{ethics}

The adoption of automatic evaluation also raises ethical concerns that warrant careful consideration. One potential issue is the possibility to continuously monitor interpreters, which might infringe on privacy rights and create a sense of constant surveillance, negatively impacting job satisfaction and professional autonomy. Additionally, decisions regarding the employability of individuals could be blindly given to mechanical means, potentially leading to unjust or biased outcomes if the algorithms fail to account for contextual nuances or other essential aspects of human communication. As such, it is crucial for the language industry to carefully weigh the benefits and challenges of automatic evaluation, ensuring that ethical considerations are addressed as advancements in AI technology continue to reshape the landscape.
\section{Conclusions}\label{Conclusions}

This study aimed to analyze the correlation between automated and human evaluations of translated content. The peculiarity of this experiment is that we focused on a specific form of translation: the simultaneous interpretation of English speeches into Spanish. This mode of translation introduces unique challenges to assessment due to the nonlinear nature of the output (in spoken translation, the differences between the source and target can be more pronounced than in written translation) and varying user expectations regarding interpretation quality. We evaluated both interpretations provided by professional interpreters and those produced by a machine interpretation system. The objective was to develop a metric reflecting interpreting quality in a manner consistent with human judgment.

The direct prompting of GPT-3.5 for quality estimation on a Likert scale exhibits the highest median correlation with human evaluation. This finding establishes GPT-3.5 as the most promising tool among the evaluated methods to gauge translation quality, both for interpretations produced by humans and machines. GPT-3.5 benefits from a larger context, performing better with larger segment windows. This suggests that the model can capture and evaluate long dependencies more effectively.  

Contrary to expectations, GPT-3.5's correlation with human judgment is somewhat stronger for translations produced by professional interpreters than for machine-generated ones. This implies that GPT might be subtly more attuned to the linguistic nuances of human translation, even though it remains adept at evaluating speech translation. One possible explanation for this is that human interpreters often introduce subtle contextual, tonal, and idiomatic adjustments that are more aligned with GPT-3.5’s training on diverse data, whereas machine translations might adhere more strictly to equivalences. Looking forward, further research could explore deeper into the characteristics of the datasets used for training such models and their alignment with real-world interpretation tasks. 

This study presents several limitations. The observed low interrater agreement suggests potential inconsistencies in human evaluations, possibly affecting correlation values, and generability of the results. Furthermore, the limited scope of the sampled translations might not capture the full range of linguistic complexities inherent to interpretation. Future research should consider evaluations for higher window sizes. In light of GPT-3.5's performance in this study, future research might explore its ability to delineate nuances of typologies of errors rather than merely providing aggregate scores

This study is considered a preliminary attempt to test the feasibility of applying automatic metrics to evaluate inputs from both human and machine interpreters. Before these metrics can be used in production, more research needs to be conducted. 


\bibliography{eamt24}

\begin{thebibliography}{}

\bibitem[\protect\citename{Artetxe \bgroup et al.\egroup }2018]{artetxe_robust_2018}
Artetxe, Mikel, Gorka Labaka, and Eneko Agirre.
\newblock 2018.
\newblock A robust self-learning method for fully unsupervised cross-lingual mappings of word embeddings.
\newblock {\em arXiv preprint arXiv:1805.06297}.

\bibitem[\protect\citename{Banerjee and Lavie}2005]{banerjee_meteor_2005}
Banerjee, Satanjeev and Alon Lavie.
\newblock 2005.
\newblock {METEOR}: {An} automatic metric for {MT} evaluation with improved correlation with human judgments.
\newblock In {\em Proceedings of the acl workshop on intrinsic and extrinsic evaluation measures for machine translation and/or summarization}, pages 65--72.

\bibitem[\protect\citename{Becerra \bgroup et al.\egroup }2013]{becerra_quality_2013}
Becerra, Olalla~García, Macarena~Pradas Macìas, and Rafael Barranco-Droege, editors.
\newblock 2013.
\newblock {\em Quality in interpreting. 1}.
\newblock Number 120 in Interlingua. Comares, Granada.

\bibitem[\protect\citename{Brown \bgroup et al.\egroup }2020]{brown_language_2020}
Brown, Tom, Benjamin Mann, Nick Ryder, Melanie Subbiah, Jared~D. Kaplan, Prafulla Dhariwal, Arvind Neelakantan, Pranav Shyam, Girish Sastry, and Amanda Askell.
\newblock 2020.
\newblock Language models are few-shot learners.
\newblock {\em Advances in neural information processing systems}, 33:1877--1901.

\bibitem[\protect\citename{Carroll}1966]{carroll_experiment_1966}
Carroll, John~B.
\newblock 1966.
\newblock An experiment in evaluating the quality of translations.
\newblock {\em Mech. Transl. Comput. Linguistics}, 9(3-4):55--66.

\bibitem[\protect\citename{Chung}2020]{chung_automatic_2020}
Chung, H.~Y.
\newblock 2020.
\newblock Automatic evaluation of human translation: {BLEU} vs. {METEOR}.
\newblock {\em Lebende Sprachen}, 65(1):181--205.

\bibitem[\protect\citename{Clark \bgroup et al.\egroup }2019]{clark_what_2019}
Clark, Kevin, Urvashi Khandelwal, Omer Levy, and Christopher~D. Manning.
\newblock 2019.
\newblock What does bert look at? an analysis of bert's attention.
\newblock {\em arXiv preprint arXiv:1906.04341}.

\bibitem[\protect\citename{Clark \bgroup et al.\egroup }2020]{clark_electra_2020}
Clark, Kevin, Minh-Thang Luong, Quoc~V. Le, and Christopher~D. Manning.
\newblock 2020.
\newblock Electra: {Pre}-training text encoders as discriminators rather than generators.
\newblock {\em arXiv preprint arXiv:2003.10555}.

\bibitem[\protect\citename{Conneau \bgroup et al.\egroup }2017]{conneau_word_2017}
Conneau, Alexis, Guillaume Lample, Marc'Aurelio Ranzato, Ludovic Denoyer, and Hervé Jégou.
\newblock 2017.
\newblock Word translation without parallel data.
\newblock {\em arXiv preprint arXiv:1710.04087}.

\bibitem[\protect\citename{Devlin \bgroup et al.\egroup }2019]{devlin_bert_2019}
Devlin, Jacob, Ming-Wei Chang, Kenton Lee, and Kristina Toutanova.
\newblock 2019.
\newblock {BERT}: {Pre}-training of {Deep} {Bidirectional} {Transformers} for {Language} {Understanding}.
\newblock {\em arXiv:1810.04805 [cs]}, May.
\newblock arXiv: 1810.04805.

\bibitem[\protect\citename{Doddington}2002]{doddington_automatic_2002}
Doddington, George.
\newblock 2002.
\newblock Automatic evaluation of machine translation quality using n-gram co-occurrence statistics.
\newblock In {\em Proceedings of the second international conference on {Human} {Language} {Technology} {Research}}, pages 138--145.

\bibitem[\protect\citename{Fantinuoli and Prandi}2021]{fantinuoli_towards_2021}
Fantinuoli, Claudio and Bianca Prandi.
\newblock 2021.
\newblock Towards the evaluation of automatic simultaneous speech translation from a communicative perspective.
\newblock In {\em Proceedings of the 18th {International} {Conference} on {Spoken} {Language} {Translation} ({IWSLT} 2021)}, pages 245--254, Bangkok, Thailand (online), August. Association for Computational Linguistics.

\bibitem[\protect\citename{Fernandes \bgroup et al.\egroup }2023]{fernandes_devil_2023}
Fernandes, Patrick, Daniel Deutsch, Mara Finkelstein, Parker Riley, André F.~T. Martins, Graham Neubig, Ankush Garg, Jonathan~H. Clark, Markus Freitag, and Orhan Firat.
\newblock 2023.
\newblock The {Devil} is in the {Errors}: {Leveraging} {Large} {Language} {Models} for {Fine}-grained {Machine} {Translation} {Evaluation}, August.
\newblock arXiv:2308.07286 [cs].

\bibitem[\protect\citename{Gerver}1969]{gerver_effects_1969}
Gerver, David.
\newblock 1969.
\newblock The effects of source language presentation rate on the performance of simultaneous conference interpreters.
\newblock In {\em Proceedings of the 2nd {Louisville} {Conference} on rate and/or frequency controlled speech}, pages 162--184. Louisville (Kty), University of Louisville.

\bibitem[\protect\citename{Gile}1995]{gile_fidelity_1995}
Gile, Daniel.
\newblock 1995.
\newblock Fidelity assessment in consecutive interpretation: {An} experiment.
\newblock {\em Target. International Journal of Translation Studies}, 7(1):151--164.
\newblock ISBN: 0924-1884 Publisher: John Benjamins.

\bibitem[\protect\citename{Han and Lu}2021a]{han_can_2021}
Han, Chao and Xiaolei Lu.
\newblock 2021a.
\newblock Can automated machine translation evaluation metrics be used to assess students’ interpretation in the language learning classroom?
\newblock {\em Computer Assisted Language Learning}, pages 1--24, August.

\bibitem[\protect\citename{Han and Lu}2021b]{han_interpreting_2021}
Han, Chao and Xiaolei Lu.
\newblock 2021b.
\newblock Interpreting quality assessment re-imagined: {The} synergy between human and machine scoring.
\newblock {\em Interpreting and Society}, 1(1):70--90, September.

\bibitem[\protect\citename{Han and Zhao}2021]{han_accuracy_2021}
Han, Chao and Xiao Zhao.
\newblock 2021.
\newblock Accuracy of peer ratings on the quality of spoken-language interpreting.
\newblock {\em Assessment \& Evaluation in Higher Education}, 46(8):1299--1313, November.

\bibitem[\protect\citename{Han}2016]{han_investigating_2016}
Han, Chao.
\newblock 2016.
\newblock Investigating {Score} {Dependability} in {English}/{Chinese} {Interpreter} {Certification} {Performance} {Testing}: {A} {Generalizability} {Theory} {Approach}.
\newblock {\em Language Assessment Quarterly}, 13(3):186--201, July.

\bibitem[\protect\citename{Han}2017]{han_using_2017}
Han, Chao.
\newblock 2017.
\newblock Using analytic rating scales to assess {English}/{Chinese} bi-directional interpretation: {A} longitudinal {Rasch} analysis of scale utility and rater behavior.
\newblock {\em Linguistica Antverpiensia, New Series–Themes in Translation Studies}, 16.
\newblock ISBN: 2295-5739.

\bibitem[\protect\citename{Han}2022]{han_interpreting_2022}
Han, Chao.
\newblock 2022.
\newblock Interpreting testing and assessment: {A} state-of-the-art review.
\newblock {\em Language Testing}, 39(1):30--55, January.

\bibitem[\protect\citename{Hendy \bgroup et al.\egroup }2023]{hendy_how_2023}
Hendy, Amr, Mohamed Abdelrehim, Amr Sharaf, Vikas Raunak, Mohamed Gabr, Hitokazu Matsushita, Young~Jin Kim, Mohamed Afify, and Hany~Hassan Awadalla.
\newblock 2023.
\newblock How good are gpt models at machine translation? a comprehensive evaluation.
\newblock {\em arXiv preprint arXiv:2302.09210}.

\bibitem[\protect\citename{Howard and Ruder}2018]{howard_universal_2018}
Howard, Jeremy and Sebastian Ruder.
\newblock 2018.
\newblock Universal language model fine-tuning for text classification.
\newblock {\em arXiv preprint arXiv:1801.06146}.

\bibitem[\protect\citename{Kalina}2012]{kalina_quality_2012}
Kalina, Sylvia.
\newblock 2012.
\newblock Quality in interpreting.
\newblock {\em John Benjamins Publishing Company}, 3:134--140.

\bibitem[\protect\citename{Kocmi and Federmann}2023]{kocmi_large_2023}
Kocmi, Tom and Christian Federmann.
\newblock 2023.
\newblock Large {Language} {Models} {Are} {State}-of-the-{Art} {Evaluators} of {Translation} {Quality}, May.
\newblock arXiv:2302.14520 [cs].

\bibitem[\protect\citename{Korpal}2012]{korpalOmissionSimultaneousInterpreting2012}
Korpal, Pawel.
\newblock 2012.
\newblock Omission in simultaneous interpreting as a deliberate act.
\newblock {\em Translation Research Projects 4}, pages 103--111.

\bibitem[\protect\citename{Lee}1999a]{lee_simultaneous_1999}
Lee, Tae-Hyung.
\newblock 1999a.
\newblock Simultaneous listening and speaking in {English} into {Korean} simultaneous interpretation.
\newblock {\em Meta: journal des traducteurs/Meta: Translators' Journal}, 44(4):560--572.
\newblock ISBN: 0026-0452 Publisher: Les Presses de l'Université de Montréal.

\bibitem[\protect\citename{Lee}1999b]{lee_speech_1999}
Lee, Tae-Hyung.
\newblock 1999b.
\newblock Speech proportion and accuracy in simultaneous interpretation from {English} into {Korean}.
\newblock {\em Meta: journal des traducteurs/Meta: Translators' Journal}, 44(2):260--267.
\newblock ISBN: 0026-0452 Publisher: Les Presses de l'Université de Montréal.

\bibitem[\protect\citename{Lee}2002]{lee_ear_2002}
Lee, Tae-Hyung.
\newblock 2002.
\newblock Ear voice span in {English} into {Korean} simultaneous interpretation.
\newblock {\em Meta: Journal des traducteurs/Meta: Translators' Journal}, 47(4):596--606.
\newblock ISBN: 0026-0452 Publisher: Les Presses de l'Université de Montréal.

\bibitem[\protect\citename{Lu and Han}2022]{lu_automatic_2022}
Lu, Xiaolei and Chao Han.
\newblock 2022.
\newblock Automatic assessment of spoken-language interpreting based on machine-translation evaluation metrics: {A} multi-scenario exploratory study.
\newblock {\em Interpreting}.
\newblock ISBN: 1384-6647 Publisher: John Benjamins Publishing Company Amsterdam/Philadelphia.

\bibitem[\protect\citename{Mackintosh}1983]{mackintosh_relay_1983}
Mackintosh, J.
\newblock 1983.
\newblock {\em {RELAY} {INTERPRETATION}: {AN} {EXPLORATORY} {STUDY}. {University} of {London}}.
\newblock {Ph.D.} thesis, unpublished MA thesis.

\bibitem[\protect\citename{Mikolov \bgroup et al.\egroup }2013]{mikolov_distributed_2013}
Mikolov, Tomas, Ilya Sutskever, Kai Chen, Greg~S. Corrado, and Jeff Dean.
\newblock 2013.
\newblock Distributed representations of words and phrases and their compositionality.
\newblock {\em Advances in neural information processing systems}, 26.

\bibitem[\protect\citename{Nia and Modarresi}2019]{nia_rasch-based_2019}
Nia, Foroogh~Khorami and Ghasem Modarresi.
\newblock 2019.
\newblock A {Rasch}-based validation of the evaluation rubric for consecutive interpreting performance.
\newblock {\em Sendebar}, 30:221--244.
\newblock ISBN: 2340-2415.

\bibitem[\protect\citename{Papineni \bgroup et al.\egroup }2002]{papineni_bleu_2002}
Papineni, Kishore, Salim Roukos, Todd Ward, and Wei-Jing Zhu.
\newblock 2002.
\newblock Bleu: a method for automatic evaluation of machine translation.
\newblock In {\em Proceedings of the 40th annual meeting of the {Association} for {Computational} {Linguistics}}, pages 311--318.

\bibitem[\protect\citename{Pennington \bgroup et al.\egroup }2014]{pennington_glove_2014}
Pennington, Jeffrey, Richard Socher, and Christopher~D. Manning.
\newblock 2014.
\newblock Glove: {Global} vectors for word representation.
\newblock In {\em Proceedings of the 2014 conference on empirical methods in natural language processing ({EMNLP})}, pages 1532--1543.

\bibitem[\protect\citename{Peters \bgroup et al.\egroup }1802]{peters_deep_1802}
Peters, Matthew~E., Mark Neumann, Mohit Iyyer, Matt Gardner, Christopher Clark, Kenton Lee, and Luke Zettlemoyer.
\newblock 1802.
\newblock Deep contextualized word representations. {CoRR} abs/1802.05365 (2018).
\newblock {\em arXiv preprint arXiv:1802.05365}.

\bibitem[\protect\citename{Pöchhacker}2002]{pochhacker_quality_2002}
Pöchhacker, Franz.
\newblock 2002.
\newblock Quality {Assessment} in {Conference} and {Community} {Interpreting}.
\newblock {\em Meta}, 46(2):410--425, October.

\bibitem[\protect\citename{Radford \bgroup et al.\egroup }2018]{radford_improving_2018}
Radford, Alec, Karthik Narasimhan, Tim Salimans, and Ilya Sutskever.
\newblock 2018.
\newblock Improving language understanding by generative pre-training.
\newblock {\em arXiv preprint}.
\newblock Publisher: OpenAI.

\bibitem[\protect\citename{Raffel \bgroup et al.\egroup }2020]{raffel_exploring_2020}
Raffel, Colin, Noam Shazeer, Adam Roberts, Katherine Lee, Sharan Narang, Michael Matena, Yanqi Zhou, Wei Li, and Peter~J. Liu.
\newblock 2020.
\newblock Exploring the limits of transfer learning with a unified text-to-text transformer.
\newblock {\em The Journal of Machine Learning Research}, 21(1):5485--5551.
\newblock ISBN: 1532-4435 Publisher: JMLRORG.

\bibitem[\protect\citename{Roweis and Saul}2000]{roweis_nonlinear_2000}
Roweis, Sam~T. and Lawrence~K. Saul.
\newblock 2000.
\newblock Nonlinear dimensionality reduction by locally linear embedding.
\newblock {\em science}, 290(5500):2323--2326.
\newblock ISBN: 1095-9203 Publisher: American Association for the Advancement of Science.

\bibitem[\protect\citename{Ruder \bgroup et al.\egroup }2019]{ruder_survey_2019}
Ruder, Sebastian, Ivan Vulić, and Anders Søgaard.
\newblock 2019.
\newblock A survey of cross-lingual word embedding models.
\newblock {\em Journal of Artificial Intelligence Research}, 65:569--631.
\newblock ISBN: 1076-9757.

\bibitem[\protect\citename{Setton and Motta}2007]{setton_syntacrobatics_2007}
Setton, Robin and Manuela Motta.
\newblock 2007.
\newblock Syntacrobatics: {Quality} and reformulation in simultaneous-with-text.
\newblock {\em Interpreting}, 9(2):199--230, January.
\newblock Publisher: John Benjamins.

\bibitem[\protect\citename{Tiselius}2009]{tiselius_revisiting_2009}
Tiselius, Elisabet.
\newblock 2009.
\newblock Revisiting {Carroll}’s scales.
\newblock {\em Testing and assessment in translation and interpreting studies}, pages 95--121.
\newblock Publisher: John Benjamins Amsterdam.

\bibitem[\protect\citename{Vaswani \bgroup et al.\egroup }2017]{vaswani_attention_2017}
Vaswani, Ashish, Noam Shazeer, Niki Parmar, Jakob Uszkoreit, Llion Jones, Aidan~N. Gomez, Łukasz Kaiser, and Illia Polosukhin.
\newblock 2017.
\newblock Attention is all you need.
\newblock {\em Advances in neural information processing systems}, 30.

\bibitem[\protect\citename{Wang \bgroup et al.\egroup }2023]{wang_document-level_2023}
Wang, Longyue, Chenyang Lyu, Tianbo Ji, Zhirui Zhang, Dian Yu, Shuming Shi, and Zhaopeng Tu.
\newblock 2023.
\newblock Document-{Level} {Machine} {Translation} with {Large} {Language} {Models}, April.
\newblock arXiv:2304.02210 [cs].

\bibitem[\protect\citename{Wu \bgroup et al.\egroup }2013]{wu_analytic_2013}
Wu, Jessica, M.~H. Liu, and Cecilia Liao.
\newblock 2013.
\newblock Analytic scoring in interpretation test: {Construct} validity and the halo effect.
\newblock {\em The making of a translator: Multiple perspectives}, 277:292.
\newblock Publisher: Bookman Taipei.

\bibitem[\protect\citename{Wu \bgroup et al.\egroup }2016]{wu_googles_2016}
Wu, Yonghui, Mike Schuster, Zhifeng Chen, Quoc~V. Le, Mohammad Norouzi, Wolfgang Macherey, Maxim Krikun, Yuan Cao, Qin Gao, and Klaus Macherey.
\newblock 2016.
\newblock Google's neural machine translation system: {Bridging} the gap between human and machine translation.
\newblock {\em arXiv preprint arXiv:1609.08144}.

\bibitem[\protect\citename{Wu}2011]{wu_assessing_2011}
Wu, Shao-Chuan.
\newblock 2011.
\newblock {\em Assessing simultaneous interpreting: {A} study on test reliability and examiners’ assessment behavior}.
\newblock {PhD} {Thesis}, Newcastle University.

\bibitem[\protect\citename{Xenouleas \bgroup et al.\egroup }2019]{xenouleas_sumqe_2019}
Xenouleas, Stratos, Prodromos Malakasiotis, Marianna Apidianaki, and Ion Androutsopoulos.
\newblock 2019.
\newblock Sumqe: a bert-based summary quality estimation model.
\newblock {\em arXiv preprint arXiv:1909.00578}.

\bibitem[\protect\citename{Yang \bgroup et al.\egroup }2019]{yang_xlnet_2019}
Yang, Zhilin, Zihang Dai, Yiming Yang, Jaime Carbonell, Russ~R. Salakhutdinov, and Quoc~V. Le.
\newblock 2019.
\newblock Xlnet: {Generalized} autoregressive pretraining for language understanding.
\newblock {\em Advances in neural information processing systems}, 32.

\end{thebibliography}
\bibliographystyle{eamt24}
\end{document}